\documentclass{article}

\usepackage{times}
\usepackage{graphicx} 
\usepackage{subfigure} 

\usepackage{natbib}

\usepackage{algorithm}
\usepackage{algorithmic}

\usepackage{hyperref}

\usepackage{amsmath,amsfonts,amssymb}

\usepackage{xspace}

\def\A{\mathcal{A}}

\def\X{\mathcal{X}}
\def\C{\mathcal{C}}
\def\Y{\mathcal{Y}}
\def\E{\mathcal{E}}
\def\P{\mathbb{P}}

\usepackage[accepted]{icml2016}

\icmltitlerunning{Separation of Concerns in Reinforcement Learning}

\begin{document} 

\twocolumn[
\icmltitle{Separation of Concerns in Reinforcement Learning}


\icmlauthor{Harm van Seijen$^{\,1}$}{harm.vanseijen@microsoft.com}
\icmlauthor{Mehdi Fatemi$^{\,1}$}{mehdi.fatemi@microsoft.com}
\icmlauthor{Joshua Romoff$^{\,12}$}{joshua.romoff@mail.mcgill.ca}
\icmlauthor{Romain Laroche$^{\,1}$}{romain.laroche@microsoft.com}
\icmladdress{$^1$Microsoft Maluuba, Montreal, Canada\\ $^2$McGill University, Montreal, Canada}


\icmlkeywords{boring formatting information, machine learning, ICML}

\vskip 0.3in
]

\begin{abstract}

In this paper, we propose a framework for solving a single-agent task by using multiple agents, each focusing on different aspects of the task. This approach has two main advantages: 1) it allows for training specialized agents on different parts of the task, and 2) it provides a new way to transfer knowledge, by transferring trained agents. Our framework generalizes the traditional hierarchical decomposition, in which, at any moment in time, a single agent has control until it has solved its particular subtask. We illustrate our framework with empirical experiments on two domains.

%

\end{abstract} 

\section{Introduction}
Marvin Minsky's "Society of Mind" theory postulates that our behaviour is not the result of a single cognitive agent, but rather the result of a society of individually simple, interacting processes called agents \citep{minsky:book88}. The power of this approach lies in specialization: different agents can have different representations, different learning processes, and so on. On a larger scale, our society as a whole validates this approach: our technological achievements are the result of many cooperating specialized agents. 

We study Minsky's idea of specialized agents in the context of reinforcement learning (RL), where the goal is to learn a policy for an agent interacting with an initially unknown environment using feedback in the form of positive or negative rewards. Specifically, we evaluate the setting of a single-agent task that is solved using multiple agents, where each agent has a different reward function. We call the resulting (non-cooperative) multi-agent system a Separation-of-Concerns (SoC) system. 

A key insight behind this work is that there is a difference between the \emph{performance objective}, which specifies what type of behaviour is desired, and the \emph{learning objective}, which is the feedback signal that modifies an agent's behaviour. In RL, a single reward function often takes on both roles. However, that these roles do not always combine well into a single function becomes clear from domains with sparse rewards, where learning can be prohibitively slow. Intrinsic motivation \citep{singh:nips04,schmidhuber:amd10} aims to address this, by adding a domain-specific intrinsic reward signal to the reward coming from the environment. Typically, the intrinsic reward function is potential-based, which maintains optimality of the resulting policy.

Our SoC system is motivated by similar ideas as intrinsic motivation, but we relax the typical assumptions to allow for more general application. Specifically, we relax assumptions along three dimensions:  1)  we do not aim for optimality of the policy, but aim to find a reasonable policy for challenging problems;  2) we allow for \emph{any} learning objective that yields good performance on the performance metric;  3) we consider multiple agents, each with a different learning objective.

This work is also related to options \citep{sutton:ai99}  and hierarchical learning \citep{dietterich:jair95, barto:deds03, kulkarni:arxiv16}. In fact, these approaches can be viewed as implementing a special type of SoC system with the agents organized in a hierarchical way; the actions of agents higher up in this hierarchy act as selectors for agents directly below it. Such a hierarchical decomposition is especially useful in sparse-reward problems where clear sub-goals can be identified. However, the more general multi-agent approach that we consider can be applied in relevant other cases as well, as we will demonstrate in this paper.

One of the main challenges in any multi-agent system is to achieve stable and independent learning (i.e., each agent learning its own value function). We distinguish different ways in which a single-agent task can be decomposed as a multi-agent SoC system and we identify for each configuration sufficient conditions for stable learning. 
Besides this analysis, we perform experiments on two domains to demonstrate the value of decomposing a task using separation of concerns.

The rest of this paper is organized as follows. We start with a motivating example to illustrate the concept of separating concerns (Section 2). After covering some basic background on MDPs (Section 3), we discuss stability conditions for different types of SoC configurations (Section 4). We then provide empirical results that illustrate the benefits and capabilities of separating concerns (Section 5). After discussing related work (Section 6), we conclude (Section 7).

\section{Motivating Example}
\label{sec:motivation}

\begin{figure}[tb]
\begin{center}
\includegraphics[width=4cm]{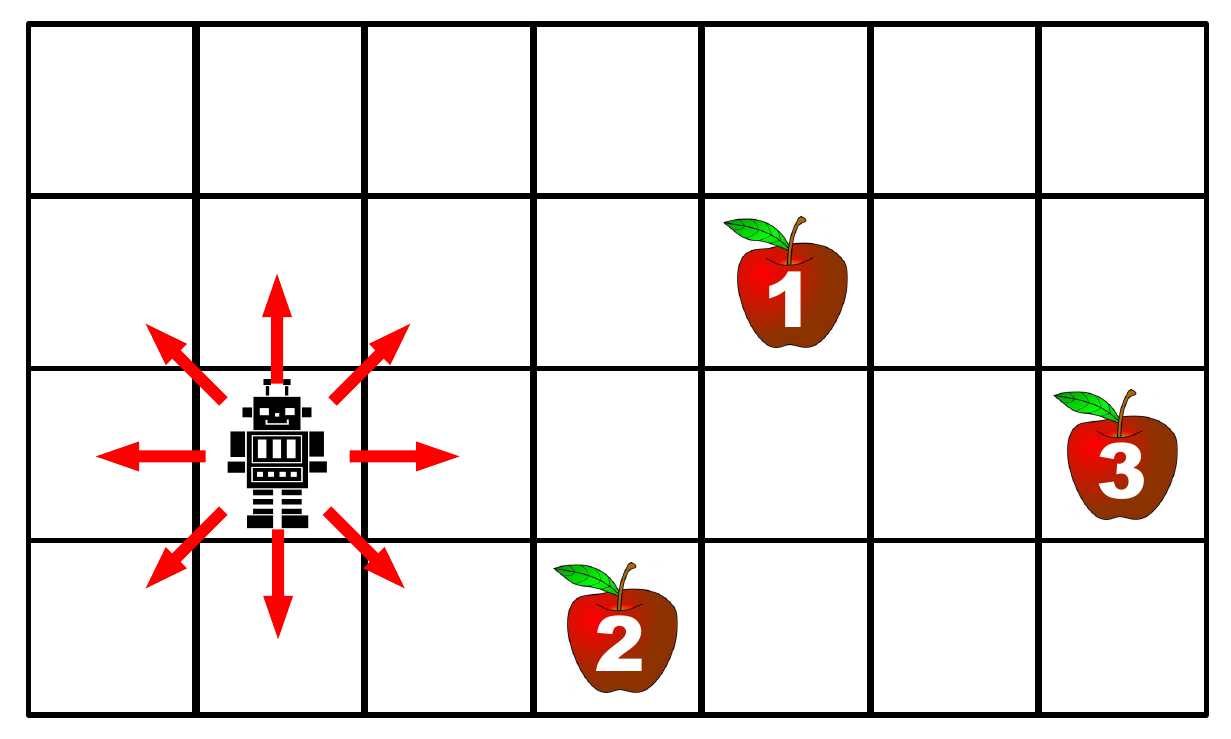}
\caption{Fruit collection example. A robot has to collect the fruits as quickly as possible (in any order). The actions are the 8 directional movements shown plus 1 no-op action. The robot receives +1 reward once all the fruits are collected; otherwise the reward is 0. The fruit positions are drawn randomly at the start of each episode. }
\label{fig:fruit collection}
\end{center}
\end{figure}

To motivate the idea of separating concerns, 
consider the fruit collection task in Figure \ref{fig:fruit collection}. The goal is to collect all the fruits as quickly as possible. 
In RL, an agent aims to maximize the \emph{return}, $G_t$, which is the expected discounted sum of rewards:
\begin{equation}
G_t = R_{t+1} + \gamma R_{t+2} + \gamma^2 R_{t+3} + \dots
\label{eq:return}
\end{equation}
By giving the agent a reward of +1 only if all the fruits are eaten, and by using a $\gamma < 1$, the optimal policy is guaranteed to use the minimal number of steps to eat all the fruits. For a grid size of 10 by 10 squares and $n$ fruits, the state-space is $100 \times 100^n  = 10^{2n + 2}$.  So for a large value of $n$, the state-space size is enormous.

Large state-spaces do not necessarily make a problem hard; by using deep reinforcement learning \citep{mnih:nature15} a task can often be mapped to some low-dimensional representation that can accurately represent the optimal value function. The problem  above, however, is an instance of the travelling salesman problem, which is known to be NP-complete \citep{papadimitriou:tcs77}. This makes it highly unlikely that some low-dimensional representation can be found that can accurately represent the optimal value function.

While the reward in the above problem is very sparse (only when \emph{all} the fruits have been eaten does the agent see a reward), this is not what makes the problem NP-complete. Adding a potential-based intrinsic reward function to make the reward less sparse will not make the problem easier, because this maintains optimality of the solution, and hence the task remains NP-complete. The task \emph{can} be made easier by adding domain knowledge in the form of a modified learning objective that 
still yields a reasonable policy with respect to the performance objective, but is easier to learn.

Consider a learning objective that gives +1 reward for eating a fruit, in combination with a $\gamma < 1$. For small $\gamma$, finding a low-dimensional representation becomes a lot easier, because fruits that are far away have minimal impact on the value function and can be ignored. A potential issue is that when all the nearby fruits are gone, the agent might not know what to do (the small values from fruits far away are likely to get drowned out by function approximation errors). On the other hand, a large $\gamma$ could be used that does not ignore fruits that are far away, but now finding a
good low-dimensional representation becomes much more challenging. 

Alternatively, consider that each fruit gets assigned a specific agent, whose only learning objective is to estimate the optimal action-value function for eating that fruit, with an aggregator making the final action selection. This agent sees a reward of +1 only if its assigned fruit gets eaten and 0 otherwise. The state-space for this agent can ignore all other fruits because they are irrelevant for its value function. Therefore, in our example, a single state-space of size $10^{2n + 2}$ gets replaced by $n$ state-spaces, each consisting of $10^4$ states. Moreover, all these $n$ agents can learn in parallel using off-policy learning. Hence, the learning problem becomes much easier. 

How well this multi-agent approach performs with respect to the performance objective (quickly eating all fruits) depends on the aggregator. The aggregator could, for example, use a voting scheme, select its action based on the summed action-values, or select its action according to the agent with the highest action-value. This last form of action selection would result in greedy behaviour, with the agent always going for the fruit that is closest, which correlates well with the performance metric. Other domains, however, might require a different aggregator.




Finally, let's consider an option-based approach. Having $n$ different fruits and one agent per fruit would now result in $n$ different options, with each option giving the policy to go to one specific fruit. These $n$ options would act as (temporally-extended) actions to a higher-level agent, which would evaluate them based on its own high-level reward function. The state-space of this higher-level agent, however, would still be the same as the flat state-space, $10^{2n+2}$, so the learning problem would not be reduced. 

\section{Background}

Throughout this paper, we indicate random variables by capital letters, functions by lowercase letters and sets by calligraphic font.

RL problems can be formalized as \emph{Markov decision processes} (MDPs), which can be described as 5-tuples of the form $\langle \mathcal{X}, \mathcal{A}, p, r, \gamma \rangle$, consisting of $\mathcal{X}$, the set of all states; $\mathcal{A}$, the set of all actions; $p(x'|x,a)$, the transition probability function, giving for each state $x \in \mathcal{X}$ and action  $a \in \mathcal{A}$  the probability of a transition to state $x' \in \mathcal{X}$ at the next step;  $r(x,a,x')$, the reward function, 
giving the expected reward for a transition from $(x,a)$ to $x'$. $\gamma$ is the discount factor, specifying how future rewards are weighted with respect to the immediate reward.  The goal is to maximize the return (Equation \ref{eq:return}) 

Actions are taken at discrete time steps $t = 0,1,2,...$ according to a \emph{policy} $\pi: \mathcal{X} \times \mathcal{A} \rightarrow [0,1]$, which defines for each action the selection probability conditioned on the state. Each policy $\pi$ has a corresponding action-value function $q_{\pi}(x, a)$, which gives the expected value of the return $G_t$, conditioned on the  state $x \in \mathcal{X}$ and action $a \in \mathcal{A}$:
$$q_{\pi}(x,a) = \mathbb{E}\{ G_t \,|\, X_t = x, A_t = a, \pi \}\thinspace.$$



\section{Separation of Concerns}
\label{sec:soc}

In this section, we discuss several agent configurations which decompose tasks in different ways. For each configuration, we discuss sufficient conditions for \emph{stable learning}, which we define below.


S
Given a learning method that converges to the optimal policy on a single-agent MDP task, applying this method independently to each of the agents of the SoC model, the overall policy of the SoC model converges to a fixed point. Moreover, this fixed point only depends on the SoC model and not on the particular learning algorithm that is being used.

We will indicate the MDP that defines the single-agent task by the tuple  $\langle \mathcal{X}^{flat}, \mathcal{A}^{flat}, p^{flat}, $ $r^{flat}, \gamma^{flat} \rangle$, and refer to the agent that tries to solve this MDP without any decomposition as the \emph{flat agent}.

Unless otherwise stated, the performance objective of the SoC model is to maximize the flat return (defined by $r^{flat}$ and $\gamma^{flat}$).

\subsection{SoC Model with Action Aggregation}
\label{sec:soc action agg}

\begin{figure}[tb]
\begin{center}
\includegraphics[width=8cm]{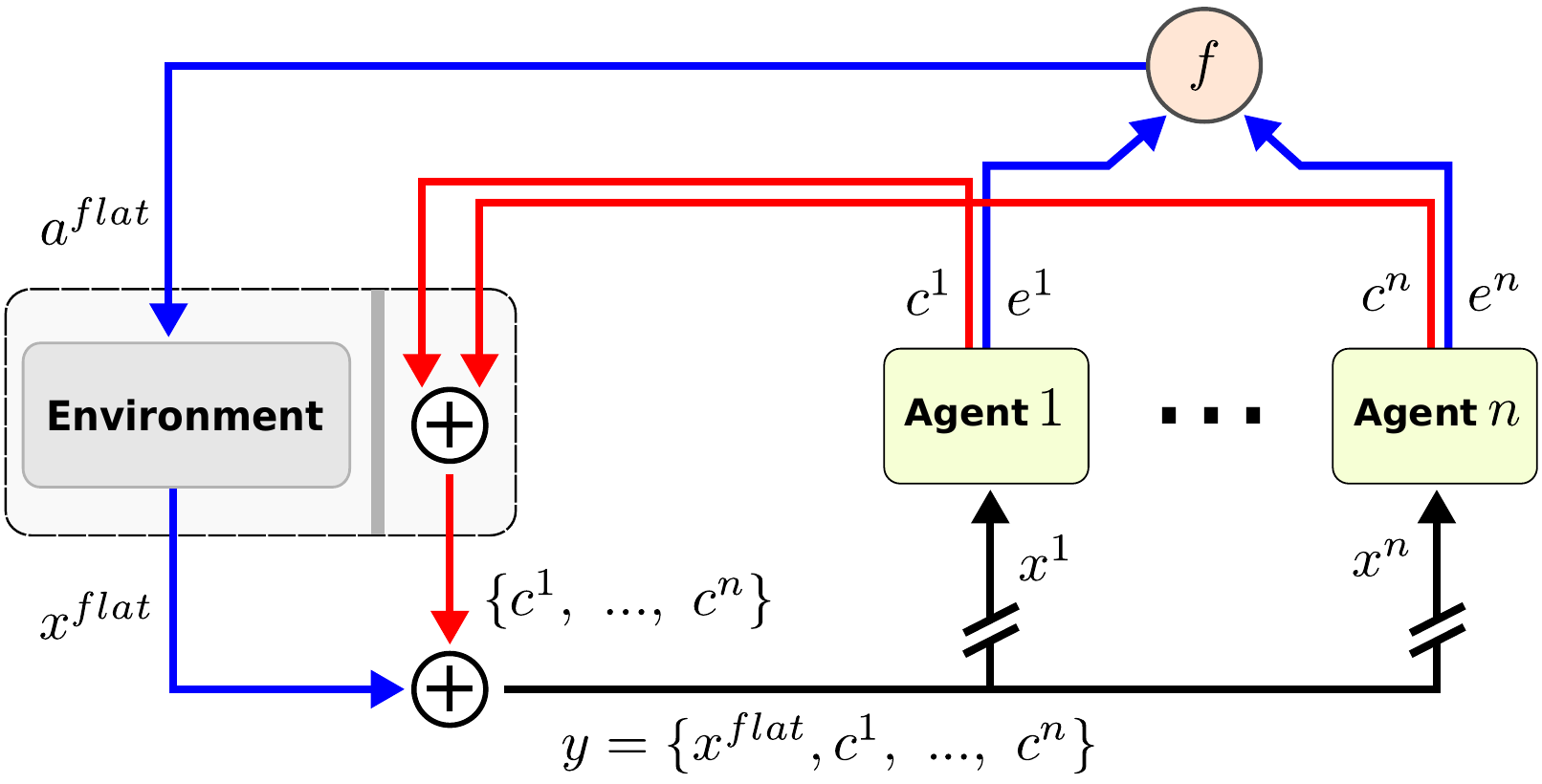}
\caption{SoC model.}
\label{fig:SoC system}
\end{center}
\end{figure}

A general way to decompose a single-agent task using $n$ agents is shown in Figure \ref{fig:SoC system}. At each time step $t$, an agent $i$ choses an action $a_t^i := (e_t^i, c_t^i) \in \A^i := \E^i \times \C^i$, with $\E^i$ its set of environment actions (which affect $\X^{flat}$), and $\C^i$ its set of communication actions (which do not affect $\X^{flat}$). We also allow for agents that only have communication actions or only environment actions. The environment actions of the agents are fed into an aggregator function $f$, which maps them to an action of $\A^{flat}$:
$$ f:  \E^1 \times \dots \times \E^n \rightarrow \A^{flat}.$$

The input space of an agent is based on the communication actions from the previous time steps and the updated flat state space.\footnote{The one time step delay of the communication actions is necessary for the general setting where all agents communicate in parallel. It  also occurs in similar architectures \citep[e.g., see][]{foerster:nips16}.}
In general, an agent will be partially observable and not see the full flat state space or all communication actions. Formally, state space $\X^i$ of agent $i$ is a projection of $\Y := \X^{flat} \times \C^1 \times \dots \times \C^n$ onto a subspace of $\Y$:
$$\X^i = \sigma^i (\Y)\,.$$

Each agent has its own reward function, $r^i : \mathcal{X}^{i} \times \mathcal{A}^{i} \times \mathcal{X}^{i}  \rightarrow \mathbb{R}$  and discount factor $\gamma^i : \mathcal{X}^{i} \times \mathcal{A}^{i} \times \mathcal{X}^{i}  \rightarrow [0, 1]$ and aims to find a policy $\pi^i: \mathcal{X}^i \times \mathcal{A}^i \rightarrow [0,1]$ that maximizes the return based on these functions. We also define $\Pi^{i}$ to be the space of all policies for agent $i$.

For stability analysis, we divide each implementation of this general SoC model into different categories. These categories are based on the relation between the different agents.

Note that by assigning a stationary policy to each of the agents, the sequence of random variables $Y_0, Y_1, Y_2, \dots$, with $Y_t \in \Y$ is a Markov chain. To formalize, let $\mu = \{ \pi^{1} \dots \pi^{n} \}$ define a set of stationary policies for all agents and $\mathcal{M} =  \Pi^1 \times \dots \times \Pi^n$ be the space of all such sets. The following holds:
$$\quad \P(Y_{t+1} | Y_t, \mu) = \P(Y_{t+1} | Y_t, Y_{t-1}, \dots, Y_0, \mu), ~ \forall \mu \in \mathcal{M}$$
Furthermore, let $\mu^{-i}$ be a set of stationary policies for all agents but $i$ and $\mathcal{M}^{-i}$ be the space of all such sets. The following relation holds for each agent $i$:
\begin{align}
\nonumber \P(X^i_{t+1} | Y_t, A^i_t, \mu^{-i}) & = \P(X^i_{t+1} | Y_t, A^i_t, ..., Y_0, A^i_0, \mu^{-i}), \\
\nonumber & \forall \mu^{-i} \in \mathcal{M}^{-i}
\end{align}
For our stability analysis, we assume the following equation to hold for all agents $i$:
\begin{equation}
\mathbb{P}(X^{i}_{t+1}  |  X^{i}_t, A_t^{i}, \mu^{-i}) = \mathbb{P}(X^{i}_{t+1}  |  Y_t, A_t^{i}, \mu^{-i}), ~ \forall \mu^{-i} \in \mathcal{M}^{-i}
\label{eq:stability condition}
\end{equation}
The interpretation of this equation is that when all agents except agent $i$ use a stationary policy, the task for agent $i$ becomes Markov.
Note that this trivially holds, if agent $i$ is not partially observable, that is, if $\X^i = \Y$.

\begin{figure}[thb]
\begin{center}
\includegraphics[width=5.5cm]{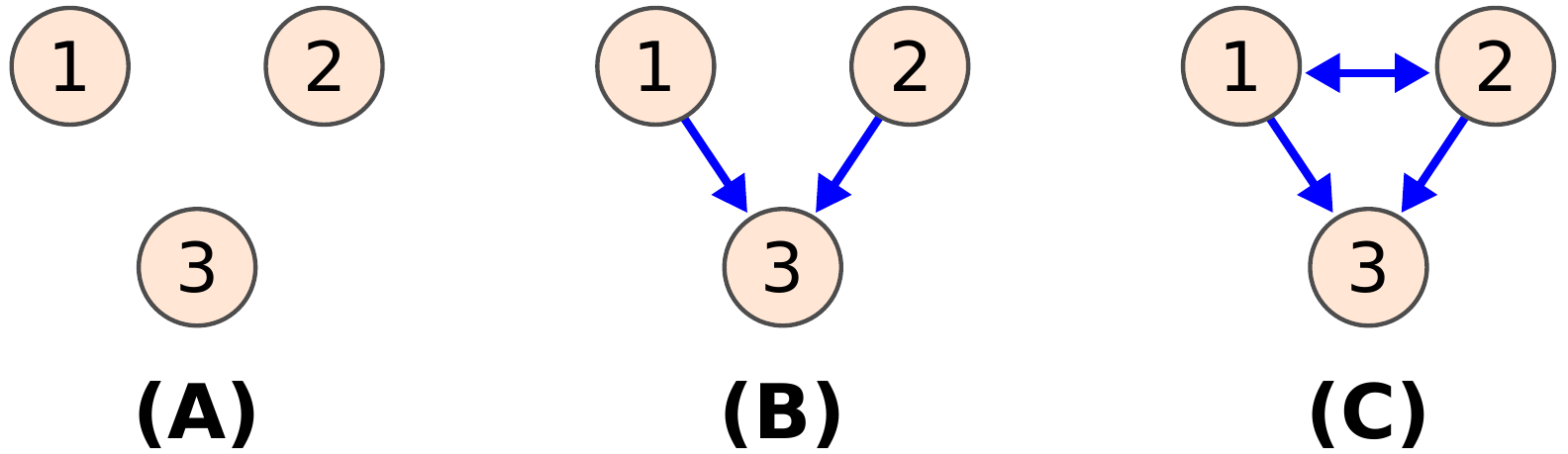}
\caption{Dependency graphs, showing the dependency relations of SoC agents (A: fully independent; B: acyclic; C: cyclic). An arrow from agent $i$ to agent $j$ means that the transition dynamics of agent $j$ depends on the policy of agent $i$.} 
\label{fig:dependency graphs}
\end{center}
\end{figure}

Under the assumption that Equation (\ref{eq:stability condition}) holds, we define agent $i$ to be \emph{independent} of agent $j$ if the policy of agent $j$ does not affect the transition dynamics of agent $i$ in any way. Formally, we extend our definitions with $\mu^{-i,-j}$ to be a set of stationary policies that assigns a policy to each agent except for agent $i$ and $j$, and $ \mathcal{M}^{-i,-j}$ to be the space of all such sets. Then, agent $i$ is independent of agent $j$ if:
\begin{align}
\nonumber \mathbb{P}(X^{i}_{t+1}  |  X^{i}_t, A_t^{i},\mu^{-i,-j},\pi^j ) &= \mathbb{P}(X^{i}_{t+1}  |  X^{i}_t, A_t^{i},\mu^{-i,-j},\hat \pi^j), \\
\forall \mu^{-i,-j} \in \mathcal{M}^{-i, -j}, &\pi^j, \hat \pi^j \in \Pi^{j}
\end{align}
Agent $i$ is \emph{dependent} of agent $j$, if it is not independent of $j$. We can show the dependency relations of the SoC agents using a dependency graph (Figure \ref{fig:dependency graphs}). Based on these relations, we can distinguish three different subclasses, which we discuss below.

\subsubsection{Independent agents}

In this case, all the agents are fully independent of each other. As an example, consider the fruit collection task from Figure \ref{fig:fruit collection} with only one fruit, at position $(fruit_{hor}, fruit_{ver})$. We can split the 9 actions of the flat agent into a horizontal action set: $A^{h} = \{\texttt{west}, \texttt{no-op}, \texttt{east}\}$ and vertical action set $A^{v} = \{\texttt{north}, \texttt{no-op}, \texttt{south}\}$, with $A^{flat} = A^{h} \times A^{v}$. The task can now be decomposed using two agents: a horizontal agent that sees state $(agent_{hor}, fruit_{hor})$ and receives a +1 reward if $agent_{hor} = fruit_{hor}$; a vertical agent that is defined similarly, but for the vertical direction.

{\bf Stability} From Equation (\ref{eq:stability condition}) and the fact that all agents are fully independent, it follows trivially that all agents converge independent of each other. Hence, stable parallel learning occurs.

\subsubsection{Acyclic Dependency}

\begin{figure}[thb]
\begin{center}
\includegraphics[width=4.0cm]{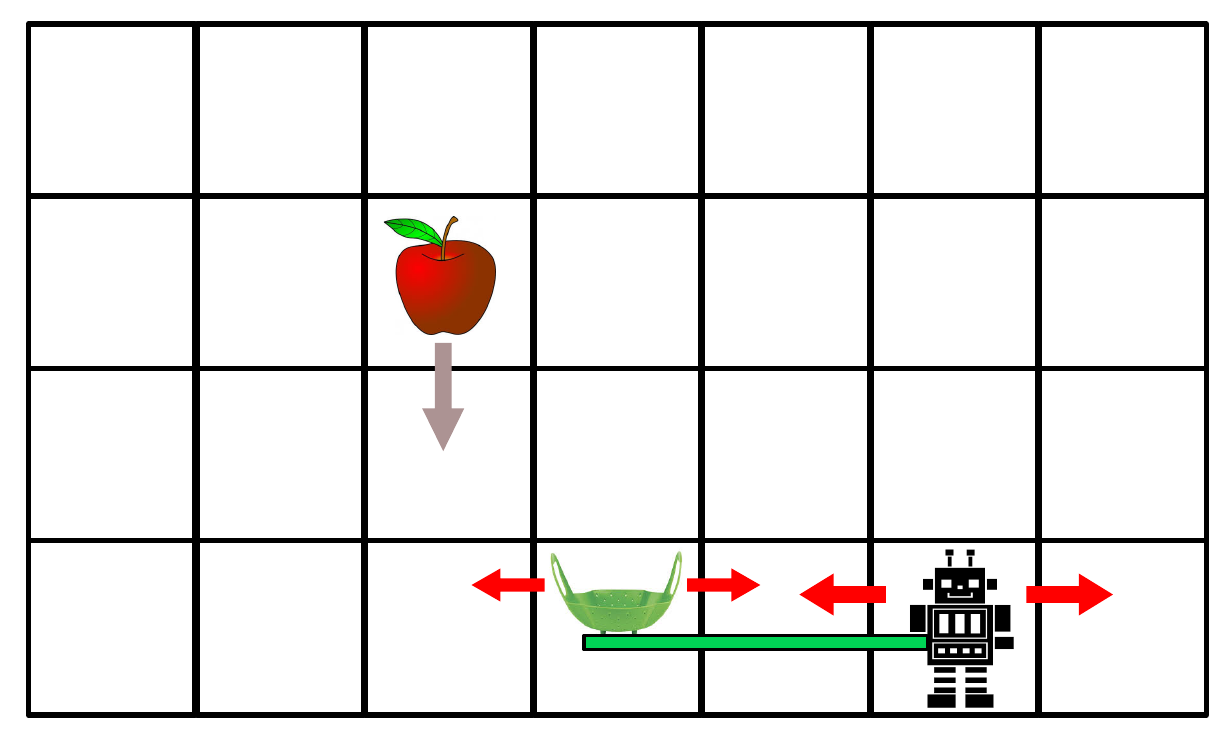}
\caption{Falling fruit example. A robot has to catch a falling fruit with the green basket to receive a reward of +1. The basket is attached to a main body via an arm, which can move relative to the body. The robot can move the main body along the horizontal axis using $\A_{body} = \{\texttt{left}, \texttt{no-op}, \texttt{right}\}$. Independent of that, the robot can move the basket (a limited amount) left or right of the body using $\A_{arm} = \{\texttt{left}, \texttt{no-op}, \texttt{right} \}$.}
\label{fig:falling fruit}
\end{center}
\end{figure}

When the dependency graph is acyclic some of the agents depend on other agents, while some of the agents are fully independent. As an example, consider the fruit catching task shown in Figure \ref{fig:falling fruit} with $A^{flat} = \A^{body} \times \A^{arm}$. 
Consider a decomposition with a `body agent' and an `arm agent'. The body agent controls $\A^{body}$, observing $(fruit_{hor}, fruit_{ver}, agent_{hor})$ and receiving +1 reward if $body_{hor} = fruit_{hor}$. The arm agent controls $\A^{arm}$, observes $(fruit_{hor}, fruit_{ver}, agent_{hor}, basket_{hor})$ and receives +1 reward if the fruit is caught. In this case, the body agent is fully independent, while the arm agent depends on the body agent.

{\bf Stability} An acyclic graph contains some fully independent agents, whose policies will converge independent of the other agents. Once these policies have converged, the agents that only depend on these independent agents will converge, and so on, until all the agents have converged. Hence, also in this case, stable parallel occurs.

\subsubsection{Cyclic Dependency}

This is the most general setting. As an example, consider again the falling fruit task, but now both agents see the full state-space. Further more, the body agent receives +1 reward if the fruit is caught, and the arm agent receives +1 reward if $basket_{hor} = fruit_{hor}$. While condition (\ref{eq:stability condition}) still holds, both agents now depend upon each other.

{\bf Stability}  For this setting, there is no guarantee of stable parallel learning, because the learning of one agent causes the environment to be non-stationary for the other agent, and vice versa. A possible approach for non-parallel learning is 
grouped coordinate descent \citep{bezdek:ota87}, which involves iteratively learning the policy of one agent, while freezing the policies of the others, and rotating which policy learns until convergence occurs \citep[see, for example,][]{thomas:icml2011}. This approach does not provide convergence in our case, however, because it requires that all agents have the same reward function. That being said, a single iteration of grouped coordinate descent (or a few) gives
a well-defined fixed point. And since we have not made statements about how close a fixed point is to the optimal policy, it is as good a fixed point as any of the other fixed points. The fixed point will depend strongly on the initial policies and the order in which the agents are updated. 

As an aside, the common approach of pre-training a low-level agent with some fixed policy, then freezing its weights and training a high-level policy using the pre-trained agent, is an instance of this more general update strategy.

\subsection{SoC for Ensemble RL}
Ensemble Learning~\cite{Dietterich2002} consists of using a large amount of weak learners in order to build a strong learner.
Applied to RL, the idea of weak learners has been abandoned for the sake of performance because of the inability of framing the RL problem into smaller problems. For example, \citet{Wiering2008} use a combination of strong RL algorithms with policy voting or value function averaging on top of it to build an even stronger algorithm. SoC enables ensemble learning in RL with weak leaners through its local state space and local reward definitions, as we outline below.

In an ensemble setting, SoC agents train their policies on the flat action space $\mathcal{A}^{flat}$ on the basis of their local state space $\mathcal{X}^i$ and their local reward function $r^i$. Contrary to Section \ref{sec:soc action agg}, they do not send their actions to the aggregator, but instead inform the aggregator of their preferences over $\mathcal{A}^{flat}$. The aggregator then selects an action based on the preferences of all agents. Any aggregator defined in \cite{Wiering2008} may be used, as well as many others: majority voting, rank voting, $Q$-value generalized means maximizer, etc. The SoC agents are trained off-policy based on the actions taken by the aggregator since it is the controller of the SoC system. We apply this configuration in Section \ref{sec:pac-boy}.

{\bf Stability}  
Given any fixed strategy of the aggregator, stable (off-policy) learning occurs if the state-space of each agent is Markov. That is, if for all agents $i$:
$$\mathbb{P}(X^i_{t+1} |  X^i_t, A_t^{flat}) = \mathbb{P}(X^i_{t+1} |  X^i_0, A_0^{flat}, \dots,  X^i_t, A_t^{flat}) \,.$$

%
%
%
%
%
%
%

\section{Experiments}

In this section, we evaluate the SoC models from Section \ref{sec:soc} empirically. We evaluate the SoC model with action aggregation on the game of Catch, focussing in particular on communication. Furthermore, we evaluate the SoC model for ensemble RL on a game inspired by Ms. Pac-Man, and show that it beats the state-of-the-art algorithms.

\subsection{Catch}

In our first example, we compare a flat agent with the SoC model on the game Catch. Catch is a simple pixel-based game introduced by \citet{mnih:nips16}. The standard game consists of a 24 by 24 screen of binary pixels in which the goal is to catch a ball that is dropped from a random location at the top of the screen with a paddle that moves along the bottom of the screen. The available actions are $\texttt{left}$, $\texttt{no-op}$ and $\texttt{right}$. The agent receives +1 reward for catching the ball, -1 if the ball falls off the screen and 0 otherwise. 

We performed experiments on the standard game with a screen size of 24 by 24, as well as a scaled up versions with screen sizes of 48 by 48 and 84 by 84. For all game sizes both the ball and the paddle consist of just a single pixel. 

\subsubsection{SoC versus Flat Agent}

\begin{figure*}[tbh]
\begin{center}
\includegraphics[width=5.5cm]{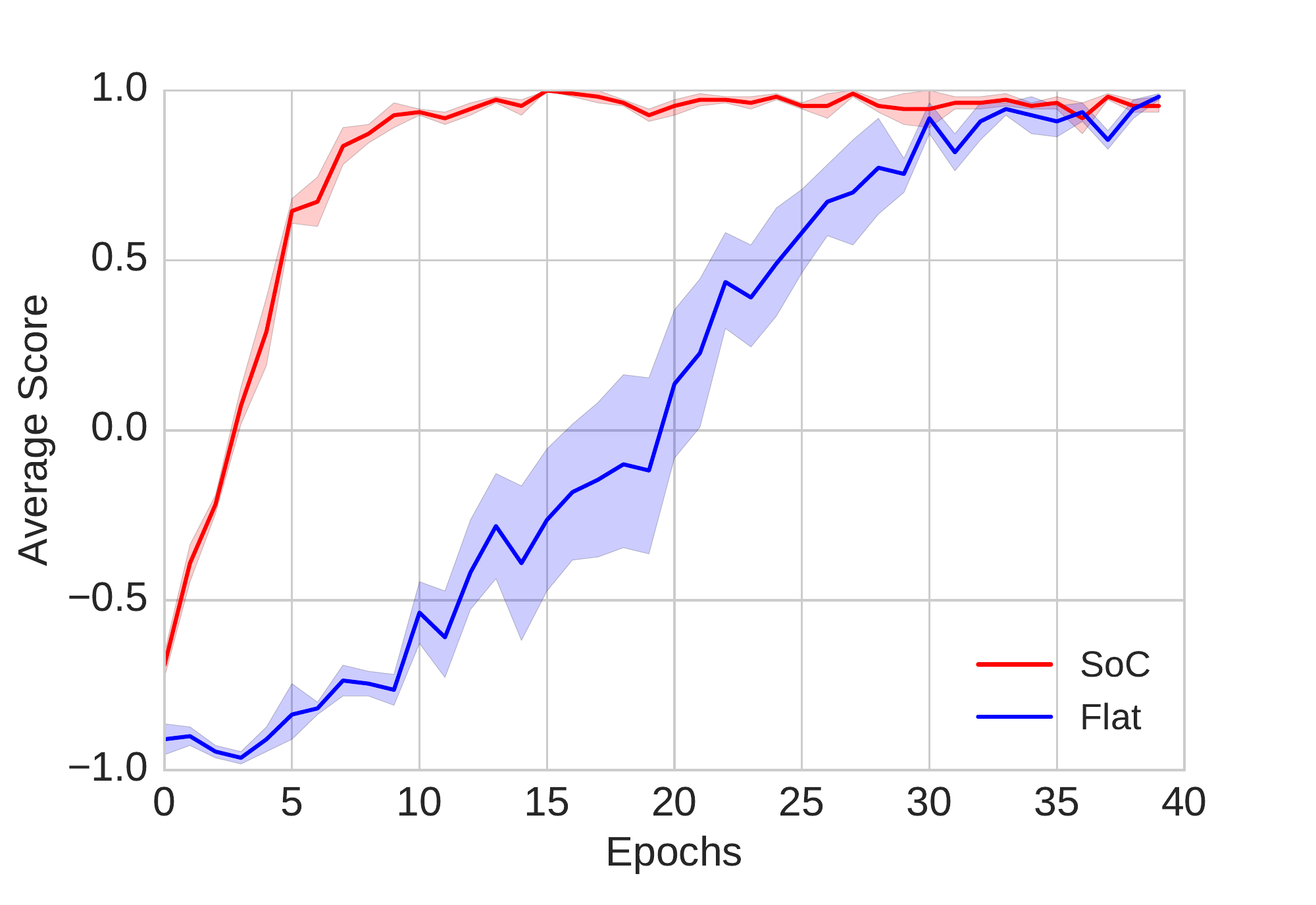}
\includegraphics[width=5.5cm]{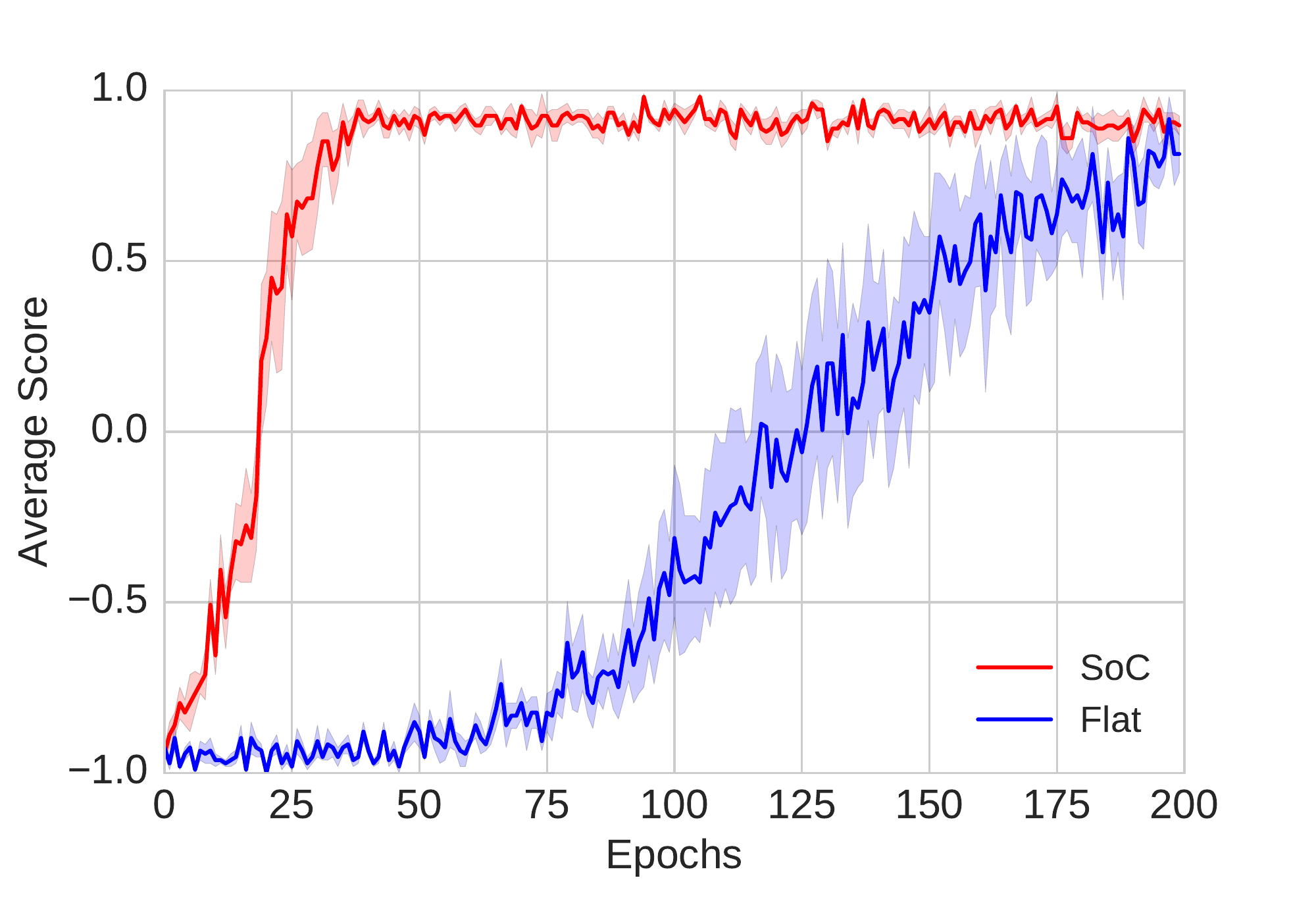}
\includegraphics[width=5.5cm]{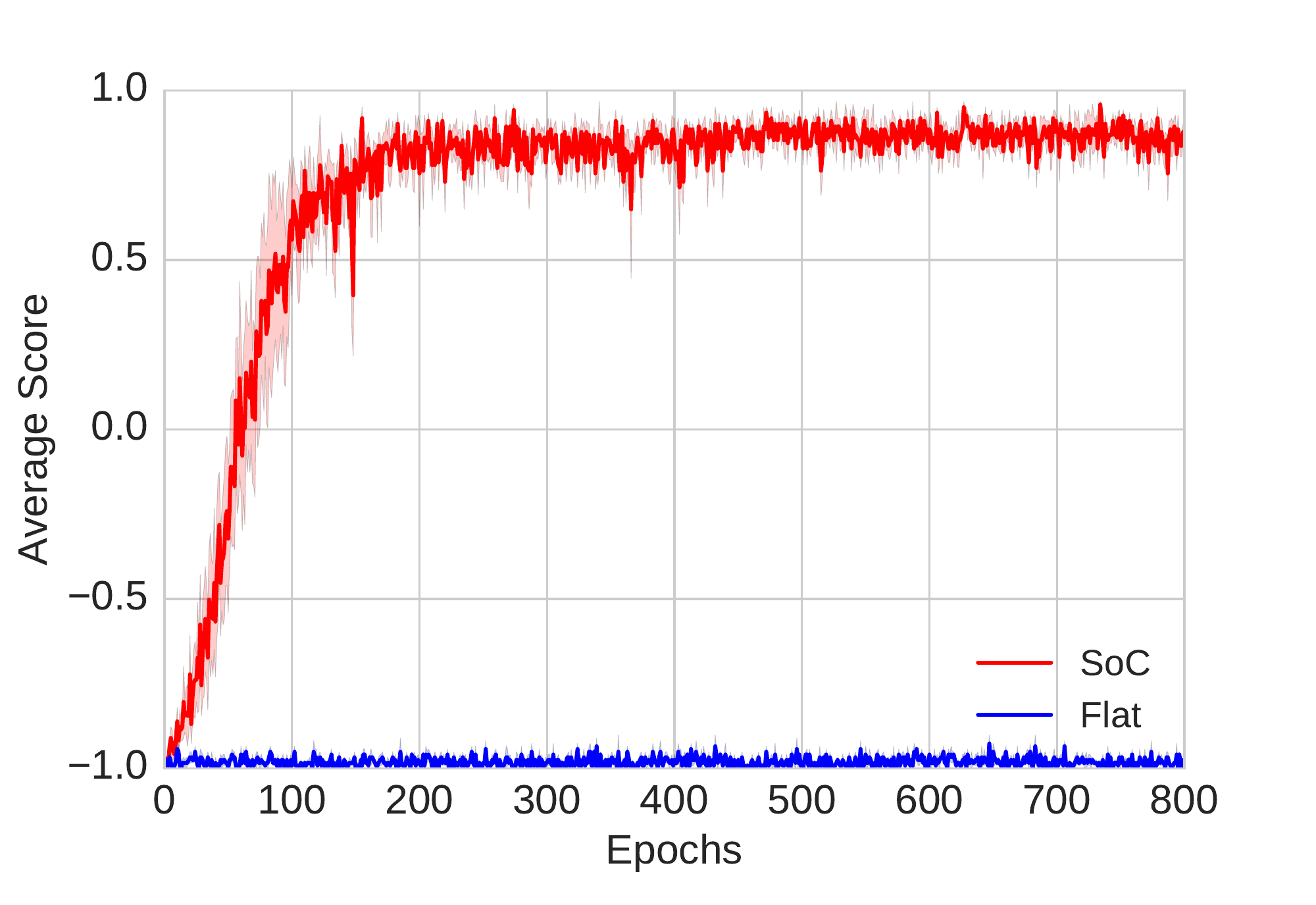}
\caption{Learning speed comparison on Catch; {\it [left]} 24x24, {\it [center]} 48x48, {\it [right]} 84x84. One epoch corresponds with 1000 actions. Each curve shows the average performance over 5 random seeds. }
\label{fig:catch}
\vspace{-0.1cm}
\end{center}
\end{figure*}

First, we compare the performance of an SoC model with action aggregation with that of the flat agent.
Our SoC model consists of a high-level and a low-level agent. The high-level agent has only communication actions, which communicate a desired action to the low-level agent: $\C^h = \{\texttt{left}, \texttt{no-op}, \texttt{right} \}$. The low-level agent has only environmental actions:  $\E^l = \A^{flat}$. Furthermore, the high-level agent uses a large discount factor (corresponding with a large horizon) and has access to the full screen, whereas the low-level agent has a small discount factor and uses a bounding box of 15 by 15 pixels around the paddle.
The high-level agent receives a reward of +1 if the ball is caught and -1 otherwise; the low-level agent receives the same reward plus a small positive reward for taking the action suggested by the high-level agent. The high-level agent takes actions every 2 time steps, whereas the low-level agent takes actions every time step. There is a cyclic dependency between the two agents. Hence, there is no guarantee of stable parallel learning. However, we found that in practice parallel learning worked well for this task.

 All agents were trained using DQN and followed a similar training/evaluation setup to \citep{mnih:nature15}. The flat agent and the high-level agent used an identical convolutional neural network. Due to the reduced state size for the low-level agent, it only requires a small dense network. 
 For the full implementation details see Appendix A and B.

The graphs in Figure \ref{fig:catch} show the results of the comparison.  The SoC model learns significantly faster than the flat agent in every tested configuration. In particular, in the 84 by 84 domain the flat agent fails to learn anything significant over the considered training period of 800 epochs. By contrast, SoC already converges after 200 epochs. The reason for the better performance of the SoC model is two-fold: the low-level agent can learn quickly due to its small state space and the high-level agent experiences a less sparse reward due to the reduced action selection frequency.

\subsubsection{Influence of the Communication Reward}

\begin{figure}[tbh]
\begin{center}
\includegraphics[width=5.5cm]{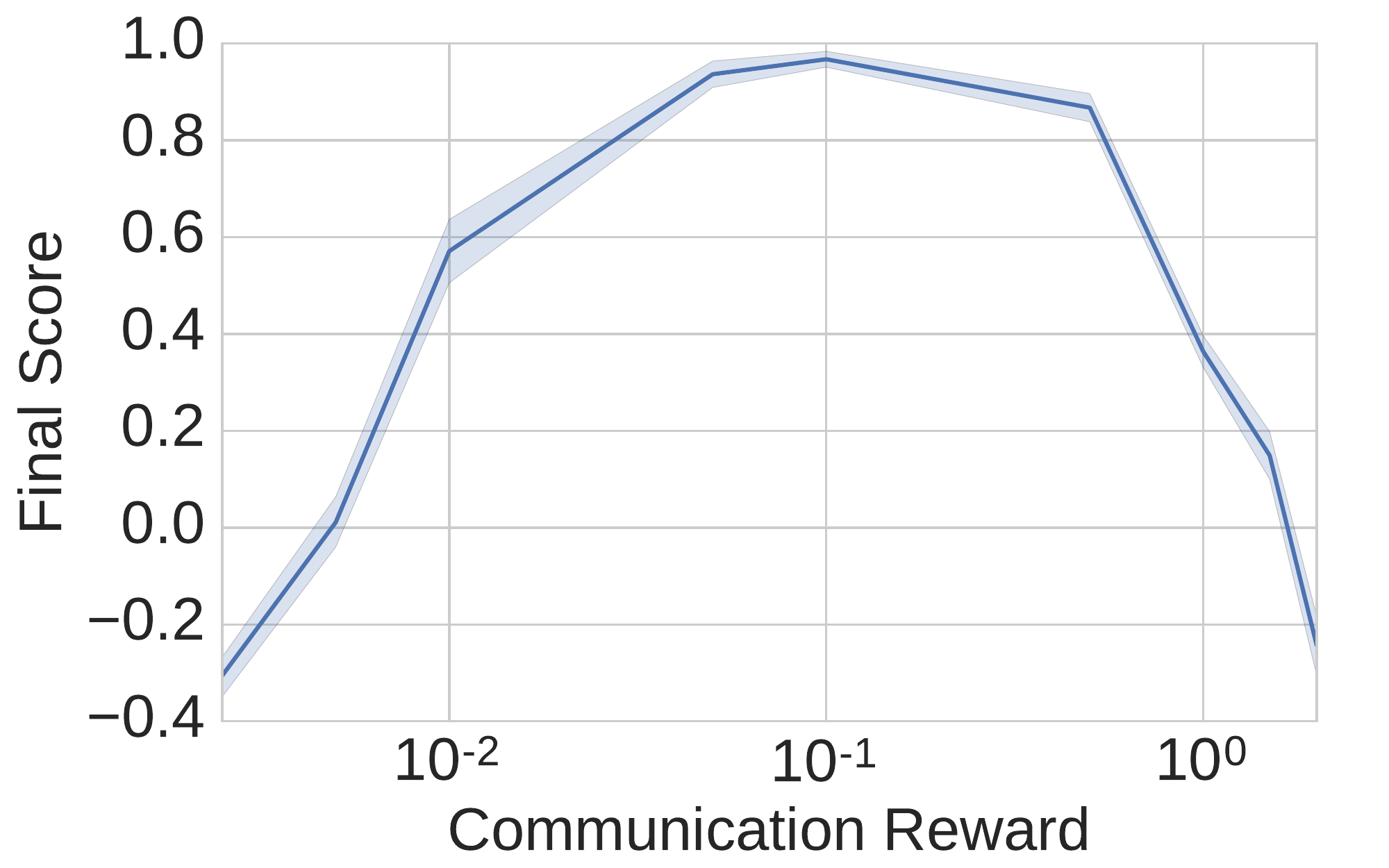}
\caption{Effect of the communication reward on the final performance of the SoC system on 24x24 Catch.}
\label{fig:catch_hill}
\end{center}
\end{figure}

To show the importance of the co-operation between the low-level and the high-level agent, we performed an additional experiment where we varied the communication reward, which is the additional reward the low-level agent receives for following the request of the high-level agent.
The results are shown in Figure \ref{fig:catch_hill}.  

\begin{figure}[t]
\begin{center}
\includegraphics[trim = 0mm 0mm 0mm 0mm, clip, width=8cm]{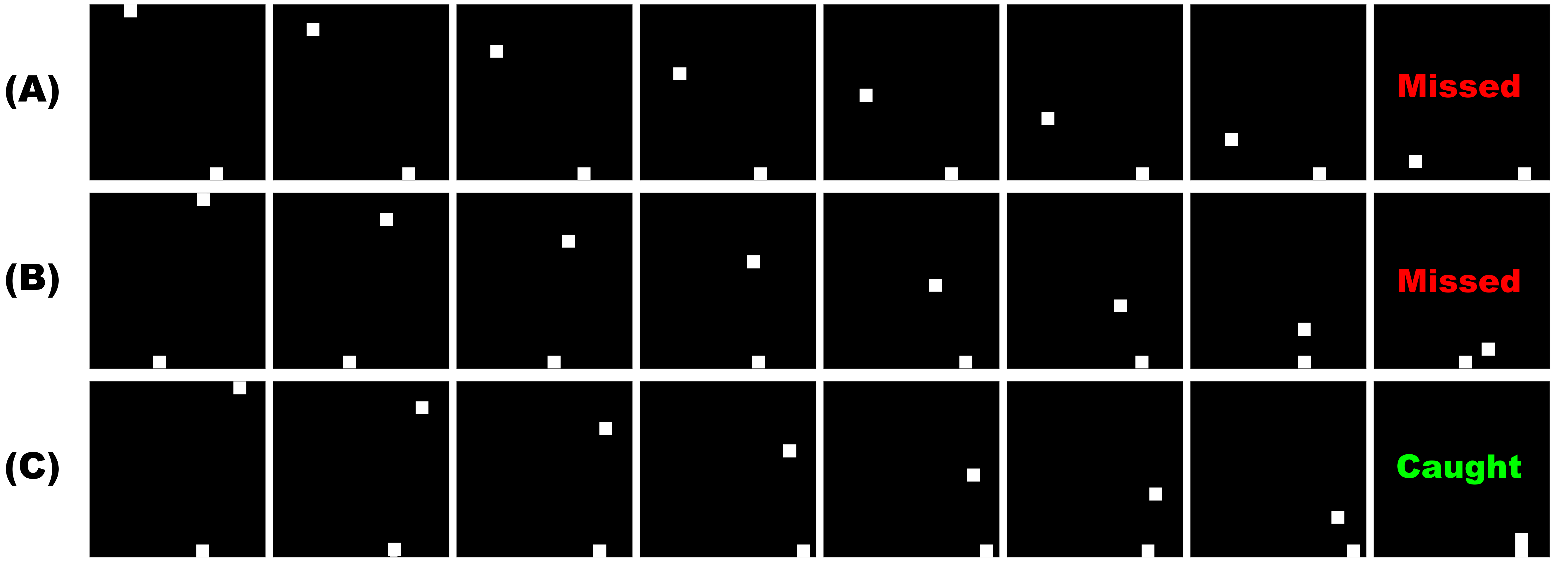}
\caption{Typical behavior on Catch for three different communication rewards; {\it (A)} reward of 0, {\it (B)} reward of 2, {\it (C)} reward of 0.1.  For clarity the ball and paddle are enhanced and we only show every 3rd frame.}
\label{fig:catch_screen}
\end{center}
\end{figure}

When the communication reward is too high or too low, the performance drops quickly. Interestingly, the reason for the performance drop is different for these two cases. This is illustrated in Figure  \ref{fig:catch_screen}, which shows a typical play for different communication rewards. If the communication reward is too low, the low-level agent ignores the requests from the high-level agent and misses balls that are dropped relatively far away from the paddle (A). If the communication reward is too high, the low-level agent will ignore the environment reward and always follow the suggestion of the high-level agent. Because the high-level agent has a low action-selection frequency, the paddle tends to overshoot the ball (B). If the communication reward is set correctly, the ball is caught almost always (C), showing that the ideal low-level agent is one that acts neither fully independent nor fully dependent.


\subsubsection{Reducing Communication}

\begin{figure}
\centering
\includegraphics[width=0.99\linewidth]{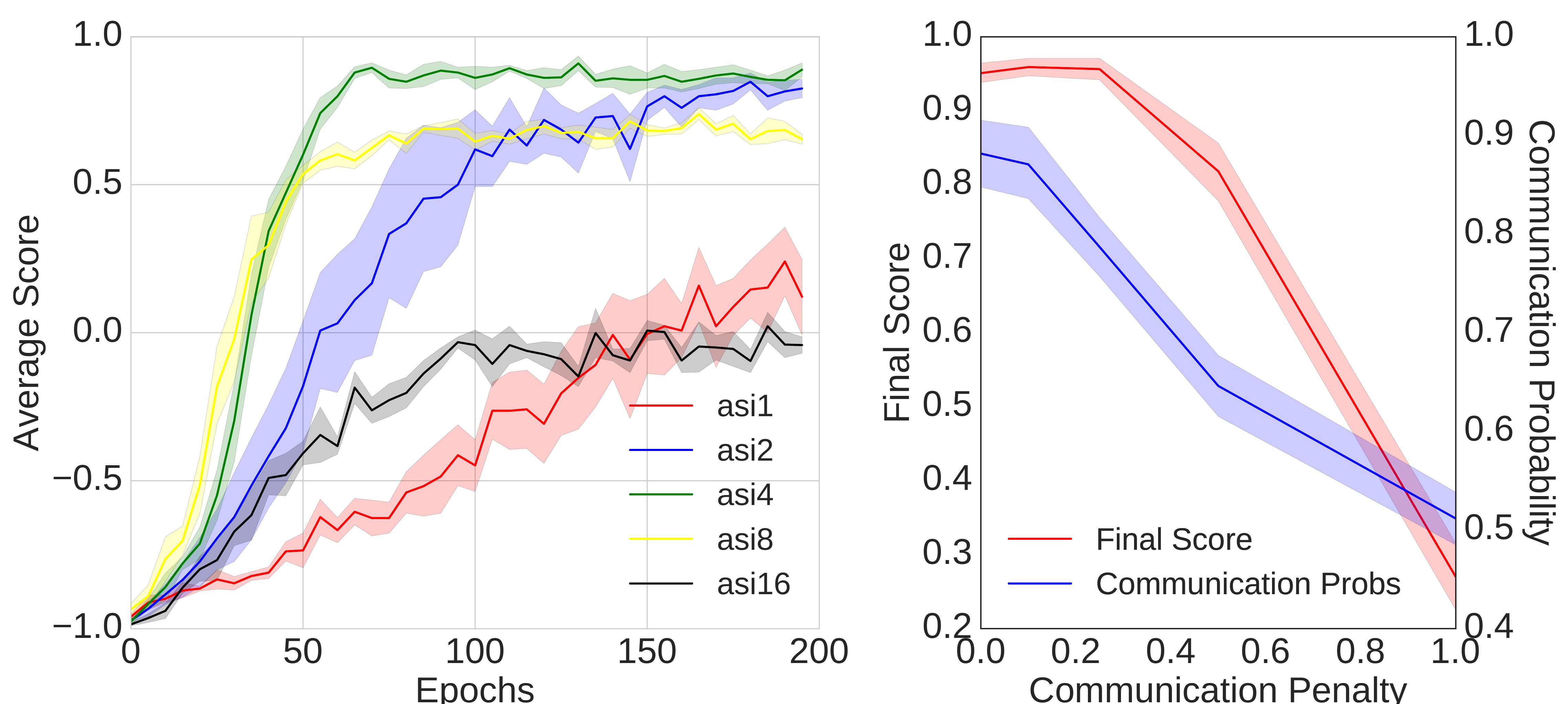}
\caption{\textit{Left:} Comparing different action selection intervals (asi) for the high-level agent of the SoC system on 84x84 Catch. \textit{Right:} Effect of penalizing communication for the high-level agent on the final performance of the SoC system on 24x24 Catch. The communication probability shows the fraction of time steps on which the high-level agent sends a communication action. }
\label{fig:catch:comm}
\end{figure}

To highlight the effect of the high-level agent's communication frequency, we tested several action-selection frequency settings in the 84 by 84 domain. The results are shown in Figure \ref{fig:catch:comm} (left).  When the communication is too frequent, the learning speed goes down, because relative to the action selections the reward appears more sparse, making learning harder. On the other hand, when it is too infrequent, asymptotic performance is reduced because the high-level agent has not enough control over the low-level agent to move it to approximately the right position.

Lastly, we tested whether the high-level agent can learn to reduce its communication on its own. To test this, we added a `no-op' action to the communication action set of the high-level agent, which does not affect the reward function of the low-level agent in any way. Furthermore, we give the high-level agent a small penalty for choosing any communication action, other than the no-op action. The action-selection frequency 
of the high-level agent is set to 1. Figure \ref{fig:catch:comm} (right) shows the results for different values of the communication penalty.
What we see is that the system can learn to maintain near optimal performance without the need for constant communication.

\subsection{Pac-Boy}
\label{sec:pac-boy}

\begin{figure}[t]
\begin{center}
\includegraphics[width=3.5cm]{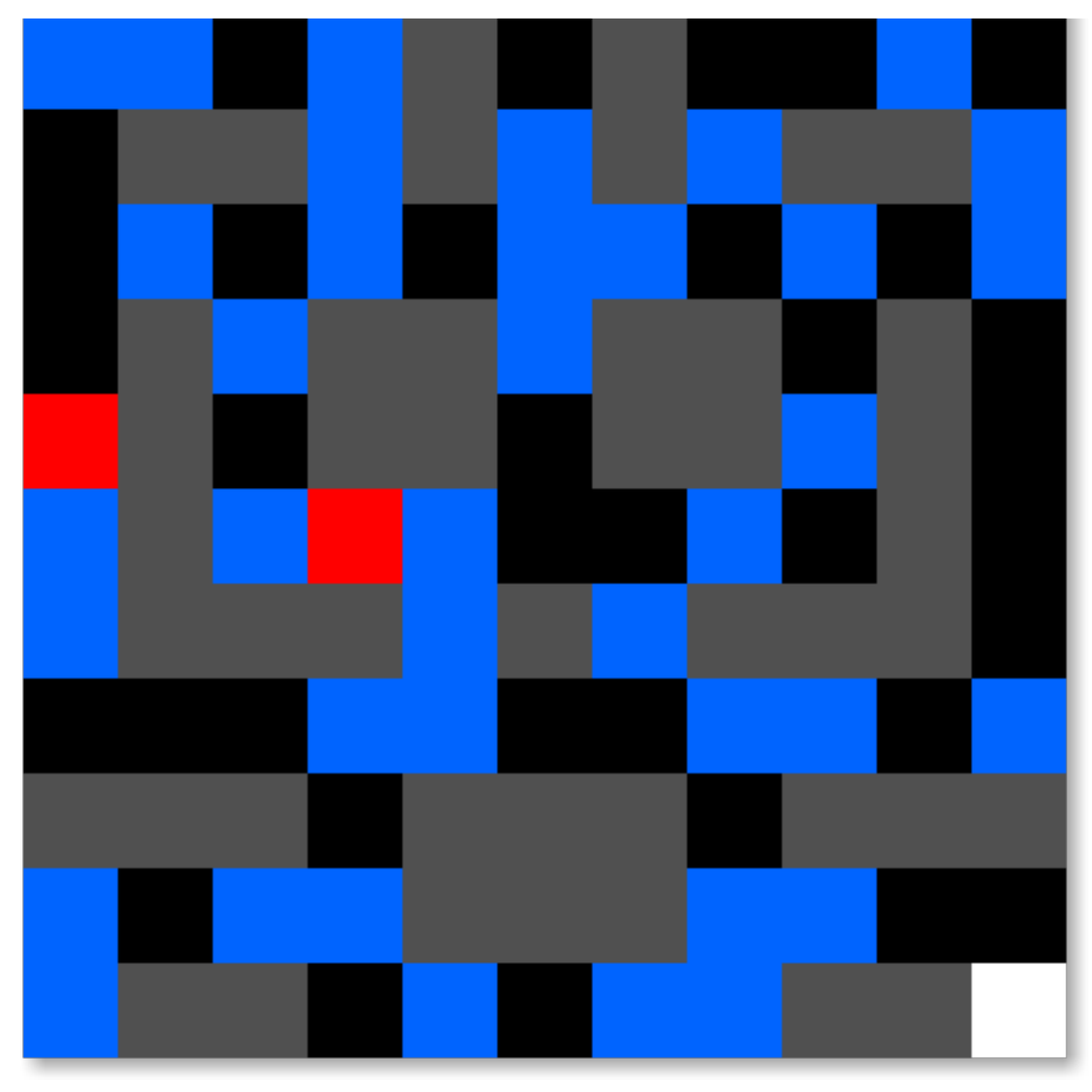}
\caption{The Pac-Boy game. Pac-Boy (white) receives a reward of +1 for eating a fruit (blue), while it gets a reward of $-10$ for bumping in one of the ghosts (red), which move randomly through the maze (walls are grey). An episode ends after all fruits are eaten or after 300 steps, whichever comes first.}
\label{fig:pacboy}
\end{center}
\end{figure}

For our next example, we evaluate an SoC model for ensemble learning on a simplified version of Ms. Pac-Man, which we call Pac-Boy (see Figure \ref{fig:pacboy}). Ms. Pac-Man is considered one of the harder games from the Atari benchmark set \citep{mnih:nature15}.

Pac-Boy contains 75 potential fruit positions. The fruit distribution is randomized. Specifically, at the start of each new episode, there is a 50\% probability for each position to have a fruit. During an episode, fruits remain fixed until they get eaten by Pac-Boy. The state of the game consists of the positions of Pac-Boy, fruits, and ghosts, resulting in $76 \times 2^{75} \times 76^{2}\approx 10^{28}$. Hence, no flat-agent can be implemented without using function approximation.

\subsubsection{SoC versus Flat Agents}
\textbf{SoC Setup} -- We separate the concerns as follows: we assign an agent to each possible fruit location. This agent sees a +1 reward only if a fruit at its assigned position gets eaten. Its state space consists of Pac-Boy's position, resulting in 76 states. 
In addition, we assign an agent to each ghost. This agent receives a -10 reward if Pac-Boy bumps into its assigned ghost. Its state space consists of Pac-Boy's position and the ghost's position, resulting in $76^2$ states. A fruit agent is only active when there is a fruit at its assigned position. Because there are on average 38 fruits, the average number of agents is 40. Due to the small state spaces of the agents, we can use a tabular representation. We train all agent in parallel with off-policy learning, using Q-learning. The aggregator function sums the Q-values for each action $a \in \mathcal{A}_{flat}$: $\,\,Q^{sum}(a, X_{t}^{flat}) :=  \sum_i Q^i(a, X^i_t)$,
and uses $\epsilon$-greedy action selection with respect to these summed values.

Interestingly, the $Q$-table of both ghost-agents are exactly the same. Hence, we can benefit from intra-task \textit{transfer learning} by sharing the $Q$-table between the two ghost-agents which results in the ghost-agents learning twice as fast. 

\textbf{Baselines} -- Our first baseline is a flat agent that uses the exact same input features as the SoC model. Specifically, the state of each agent of the SoC model is encoded with a one-hot vector and all these vectors are concatenated, resulting in a binary feature vector of size $17,252$ with about $40$ active features per time step. This vector is used for linear function approximation with Q-learning.

We then consider two Deep RL baselines. The first is the standard DQN algorithm \citep{mnih:nature15} with reward clipping. The second is Pop Art \citep{vanHasselt:nips16}, which can be combined with DQN in order to handle large magnitudes of reward (referred to as DQN-scaled).  The input to both DQN-clipped and DQN-scaled is a 4-channel binary image, where each channel is in the shape of the game grid and represents the positions of one of the following features: the walls, the ghosts, the fruits, or Pac-Boy. 
For the complete implementation details see the Supplementary document.

Figure \ref{fig:pacboy:socdqn} shows the learning speed of the SoC model compared to the baselines described above. The upper-bound line shows the maximum average score that can be obtained. What we see is that SoC converges to a policy, which is very close to optimal, whereas the baselines fall considerably short. The linear baseline must handle the massive state space with absolutely no reductions and thus takes considerably longer to converge.  While DQN-clipped and DQN-scaled converge to similar final performances, their policies differ a lot (see Steps in Figure \ref{fig:pacboy:socdqn}). DQN-scaled is much wearier of the high negative reward obtained from being eaten by the ghosts and thus takes much more time to eat all the fruit.  

\begin{figure}
\centering
\begin{subfigure}
  \centering
  \includegraphics[width=.49\linewidth]{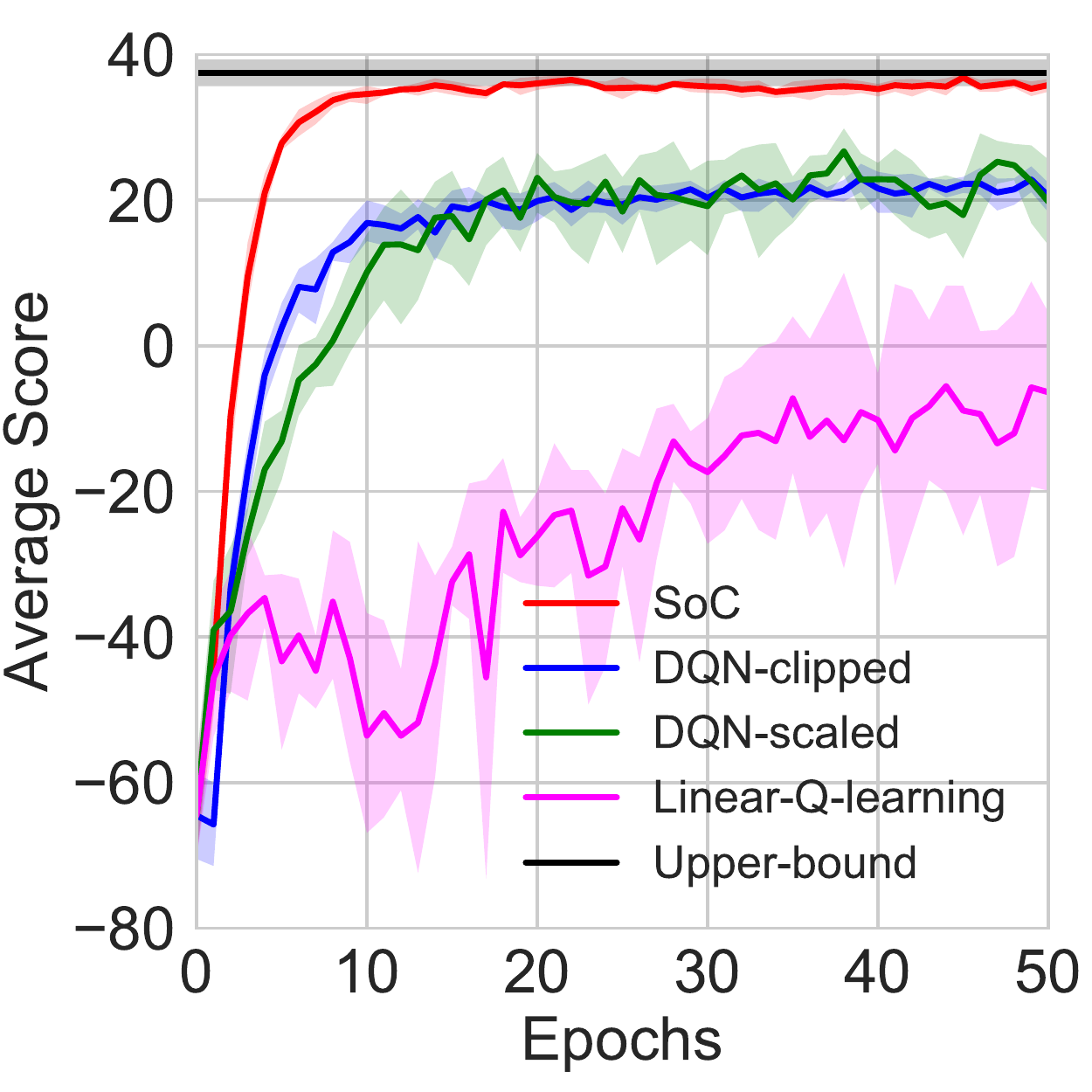}
\end{subfigure}%
\begin{subfigure}
  \centering
  \includegraphics[width=.49\linewidth]{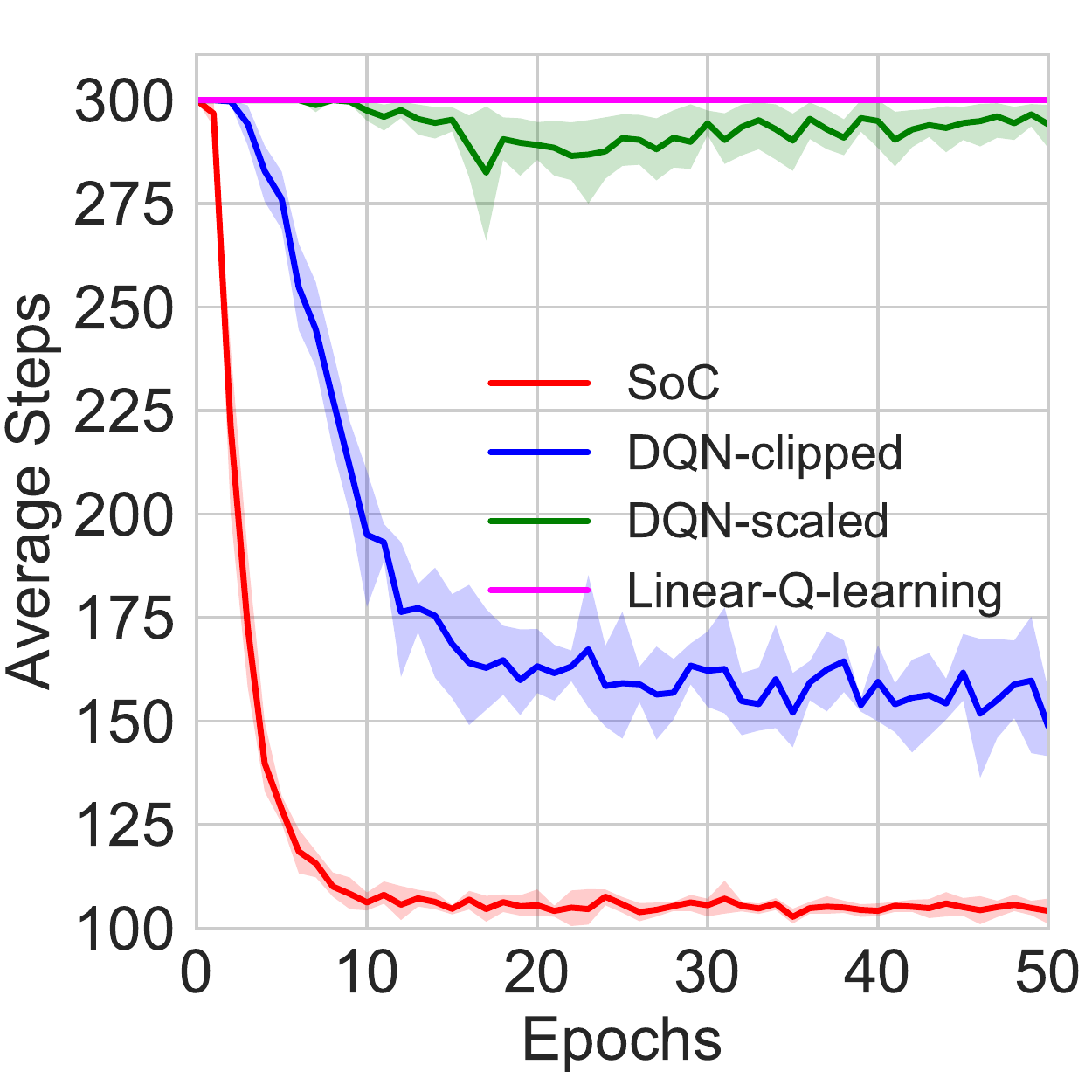}
\end{subfigure}
\caption{Learning speed comparison on Pac-Boy. One epoch corresponds with 20k environmental steps and each curve shows the average performance over 5 random seeds.}
\label{fig:pacboy:socdqn}
\end{figure}

\subsubsection{Transfer Learning}

\begin{figure}
\centering
\begin{subfigure}
  \centering
  \includegraphics[width=.49\linewidth]{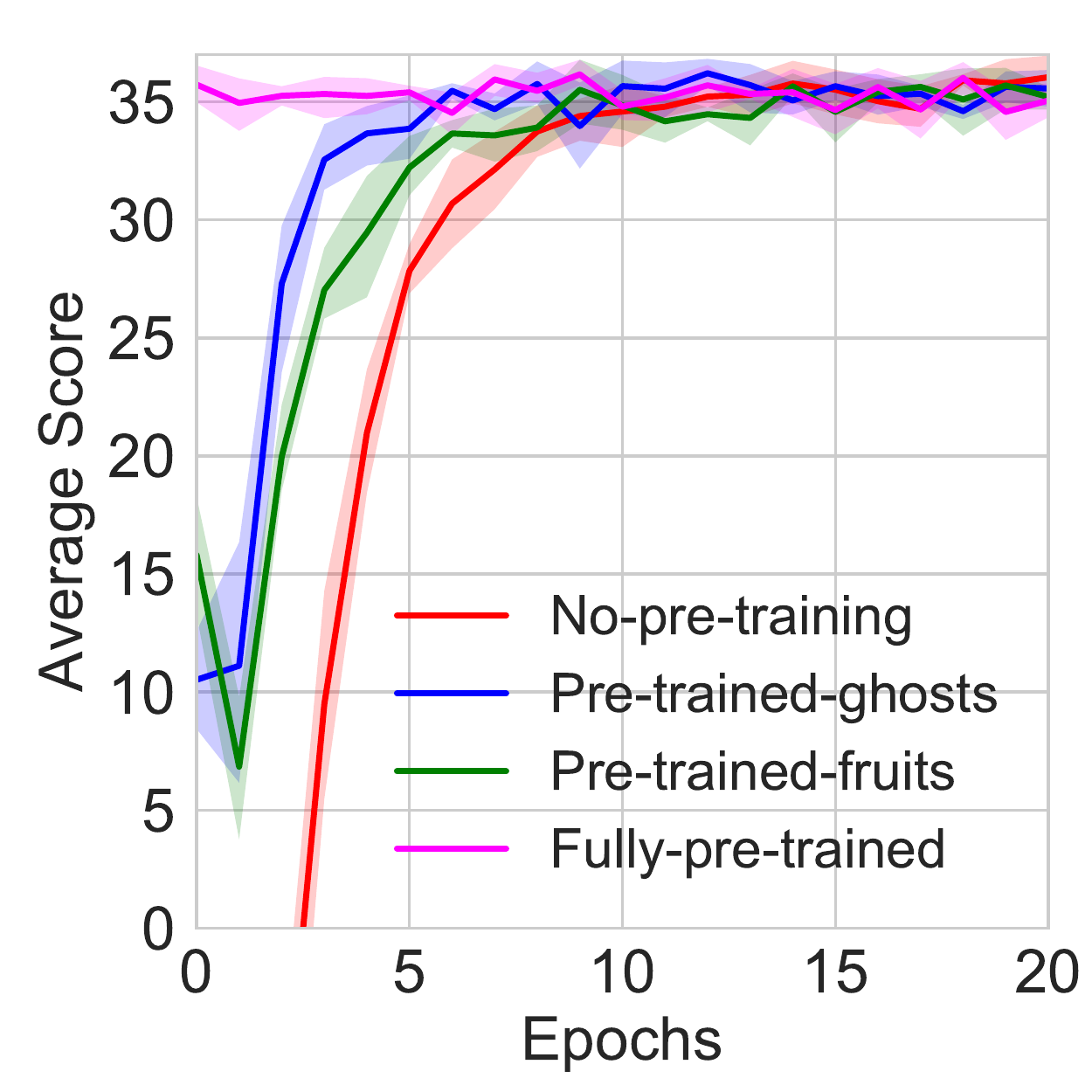}
\end{subfigure}%
\begin{subfigure}
  \centering
  \includegraphics[width=.49\linewidth]{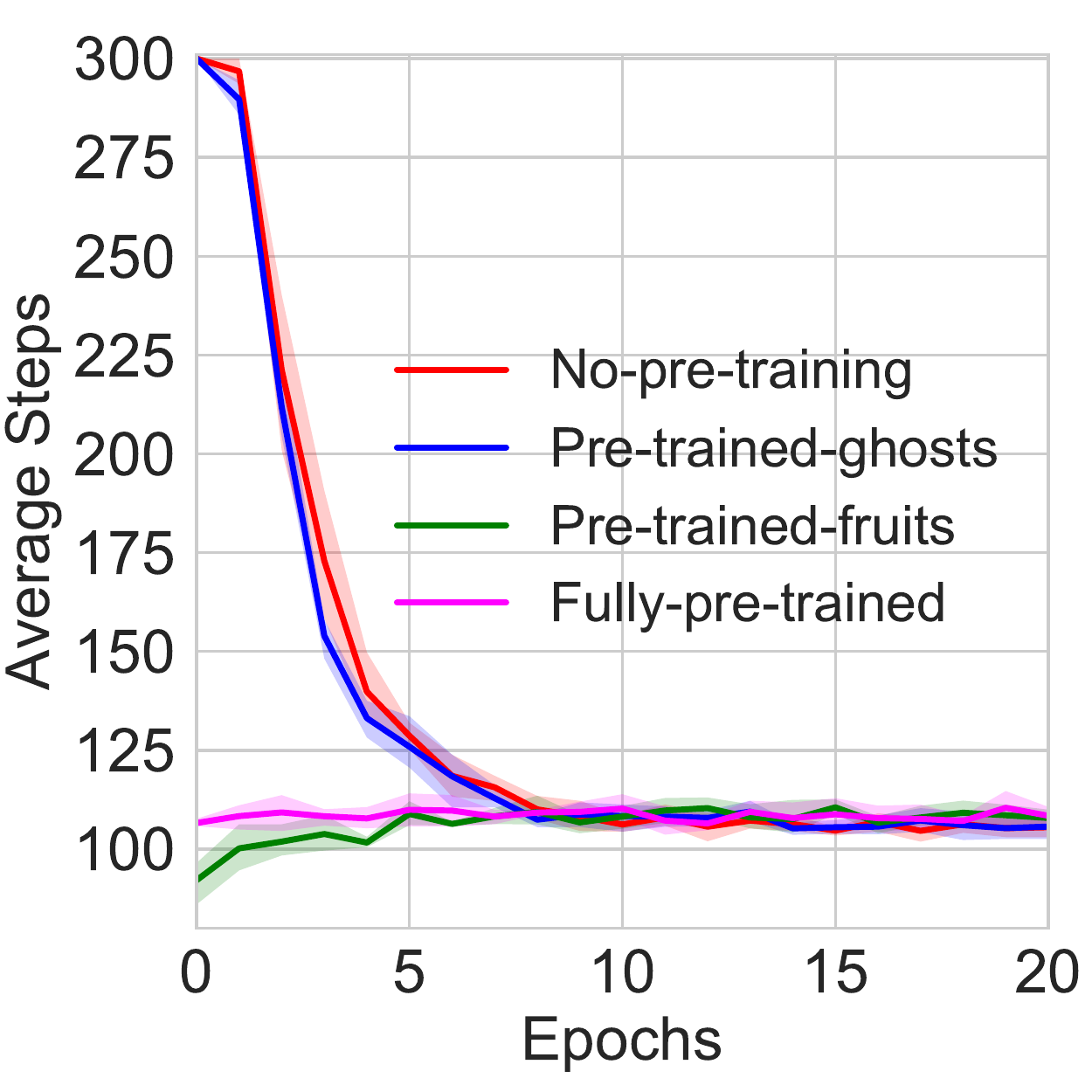}
\end{subfigure}
\caption{SoC with and without pre-training on Pac-Boy.}
\label{fig:pacboy:socpre}
\end{figure}

In order to evaluate SoC's capability for knowledge-transfer, we tested different forms of pre-training. Specifically, we compared the following: 1) pre-trained ghost agents,  2) pre-trained fruit agents, and 3) (separately) pre-trained fruit and ghost agents. We perform pre-training using a random behaviour policy. After pre-training, the agents are transferred to the full game and the remaining agents are trained. As can be seen in Figure \ref{fig:pacboy:socpre}, the knowledge transfer results in a clear boost at the beginning compared to the original SoC implementation. 

\section{Related Work}

\citet{kulkarni:arxiv16} study hierarchical RL in the context of deep reinforcement learning. In their setting, a high-level controller specifies a goal for the low-level controller. Once the goal is accomplished, the top-level controller selects a new goal for the low-level controller. The system is trained in two phases: in the first phase the low-level controller is trained on a set of different goals;  in the second phase the high-level and low-level controllers are trained in parallel. \citet{heess:arxiv16} also use a system with a high-level and a low-level controller, but the high-level controller continuously sends a modulation signal to the low-level controller, affecting its policy. This type of hierarchical learning can be viewed as a special case of an SoC system, where the agents are organized in a hierarchical way.

The work on conjugate MDPs \citep{thomas:icml2011, thomas:nips2011} is also closely related. Here, several agents coordinate to produce an action. The whole network can be trained using policy gradient. The difference with our work is that this approach can be viewed as a fully-cooperative multi-agent system (all agents share the same learning objective), whereas we consider the non-cooperative setting.

\citet{foerster:nips16} used a framework of communicating agents based on deep neural networks to solve various complex tasks. 
But like the work on conjugate MDPs, they only considered the cooperative multi-agent setting. SoC, by allowing to define different rewards for each agents, has a wider range of expressivity. We believe that the Pac-Boy experiment is a good illustration of how powerful a system can be made of non-cooperative agents.

\section{Discussion and Future Work}

We evaluated separating concerns for a single-agent task both analytically, by determining conditions for stable learning, as well as empirically, through evaluation on two domains.

We demonstrated that by giving agents a reward function that depends on the communication actions of other agents, it can be made to listen to requests from other agents to different degrees. How well it listens depends on the specific reward function. In general, agents can be made to fully ignore other agents, fully be controlled by other agents or something in between, where it makes a trade-off between following the request of another agent and ignoring it. Moreover, we showed that an agent  that retains some level of independence can in some cases yield the best overall performance. Furthermore, we demonstrated that an SoC model can convincingly beat (singe-agent) state-of-art methods on a challenging domain.

An SoC model has in common with its closest related work---intrinsic motivation and hierarchical learning---that it uses domain-specific knowledge to improve performance. There is a long line of work that aims to learn such knowledge, especially in the context of options. Although significant progress has been made  \citep[e.g., ][]{bacon:aaai17}, this remains a challenging problem. 
For our purposes, the use of domain knowledge is not a big obstacle. We aim to scale up RL such that it can be applied in specific real-world systems, for example complex dialogue systems or bot environments. In this context, using domain knowledge to achieve good performance on an otherwise intractable domain is acceptable.

In this article, we illustrated SoC on two specific settings, that we called action aggregation, and ensemble RL, but we believe that SoC's expressive power is wider and that other SoC settings are still to be discovered and evaluated.


\appendix

\begin{table*}[thb!]
\caption{Hyper-parameters used for all agents}
\label{hyper-params}
\begin{tabular}{|l|l|l|l|l|}
\hline
                           & Catch (SoC and DQN) & Pac-Boy (DQN Baselines) & Pac-Boy (SoC) & Pac-Boy (linear) \\ \hline
training steps per epoch   & 1000                & 20000  & 20000 & 20000                 \\ \hline
evaluation steps per epoch & 1000                & 10000    & 10000	& 10000                 \\ \hline                        
minibatch size             & 32                  & 32      & N/A & N/A                  \\ \hline
experience replay size     & 10000               & 100000   & N/A & N/A                \\ \hline
learning frequency         & 4                   & 1        & 1 & 1                 \\ \hline
target update frequency    & 100                 & 1000     & N/A & N/A                 \\ \hline
gamma                      & .99                 & .9       & 0.4 & 0.9                 \\ \hline
learning rate              & 0.001               & 0.00025  & 1 fruit / 0.1 ghosts & 0.005              \\ \hline
momentum                   & 0.95                & 0.95     & N/A &  N/A               \\ \hline
initial epsilon            & 1                   & 1        & 0.1 &  1               \\ \hline
final epsilon              & 0.01                & 0.1      & 0.1 &  0              \\ \hline
epsilon annealing steps    & 10000               & 100000   & 0 &  150000               \\ \hline
$\beta$                    & N/A                 & 0.00025 (Pop-Art) & N/A & N/A  \\ \hline
\end{tabular}
\end{table*}

\section{Experimental Setup}
In order to speed up learning and take advantage of these smaller domains, we tuned the parameters originally reported in Mnih et al. (2015) based on a rough search on each domain. Specifically we reduced the replay memory size, the target network update frequency, and number of annealing steps for exploration. We then did a coarse search over learning rates sampled from [0.0001, 0.00025, 0.0005, 0.00075, 0.001, 0.0025] on DQN for 24 by 24 Catch and Pac-Boy. For Pop-Art we set the learning rate to be 0.00025 (which was found to be the best learning rate for DQN on Pac-Boy) and then ran a search for the adaptive-normalization rate $\beta$ by searching over the same parameters mentioned above.  The settings that we used for all agents and experiments can be seen in Table \ref{hyper-params}.

\begin{table}[]
\caption{Filter Shapes and Strides used for DQN agents}
\label{filters-strides}
\begin{tabular}{|l|l|l|l|l|}
\hline
              & Catch  & Catch & Catch & Pac-Boy\\
              &  24x24 &  48x48 &  84x84 &  \\ \hline
Conv 1 Filter & (5, 5)      & (5, 5)      & (8, 8)      & (3, 3) \\ \hline
Conv 2 Filter & (5, 5)      & (5, 5)      & (4, 4)      & (3, 3) \\ \hline
Conv 1 Stride & (2, 2)      & (2, 2)      & (4, 4)      & (1, 1) \\ \hline
Conv 2 Stride & (2, 2)      & (2, 2)      & (2, 2)      & (1, 1) \\ \hline
\end{tabular}
\end{table}

\section{Network Architectures}
We used a core network architecture across DQN agents. The network begins by passing the input through two convolutional layers sequentially with 16 and 32 filters respectively. This is followed by two densely connected layers of size 256 and $\vert Actions \vert$. All layers except for the output use a rectified non-linear activation, whereas the output layer uses a linear activation. Depending on the domain size we vary the size of the filters and the stride for the convolutional layers, which can be seen in Table \ref{filters-strides}.

\begin{figure}[]
\begin{center}
\includegraphics[trim = 0mm 0mm 0mm 170mm, clip, width=7cm]{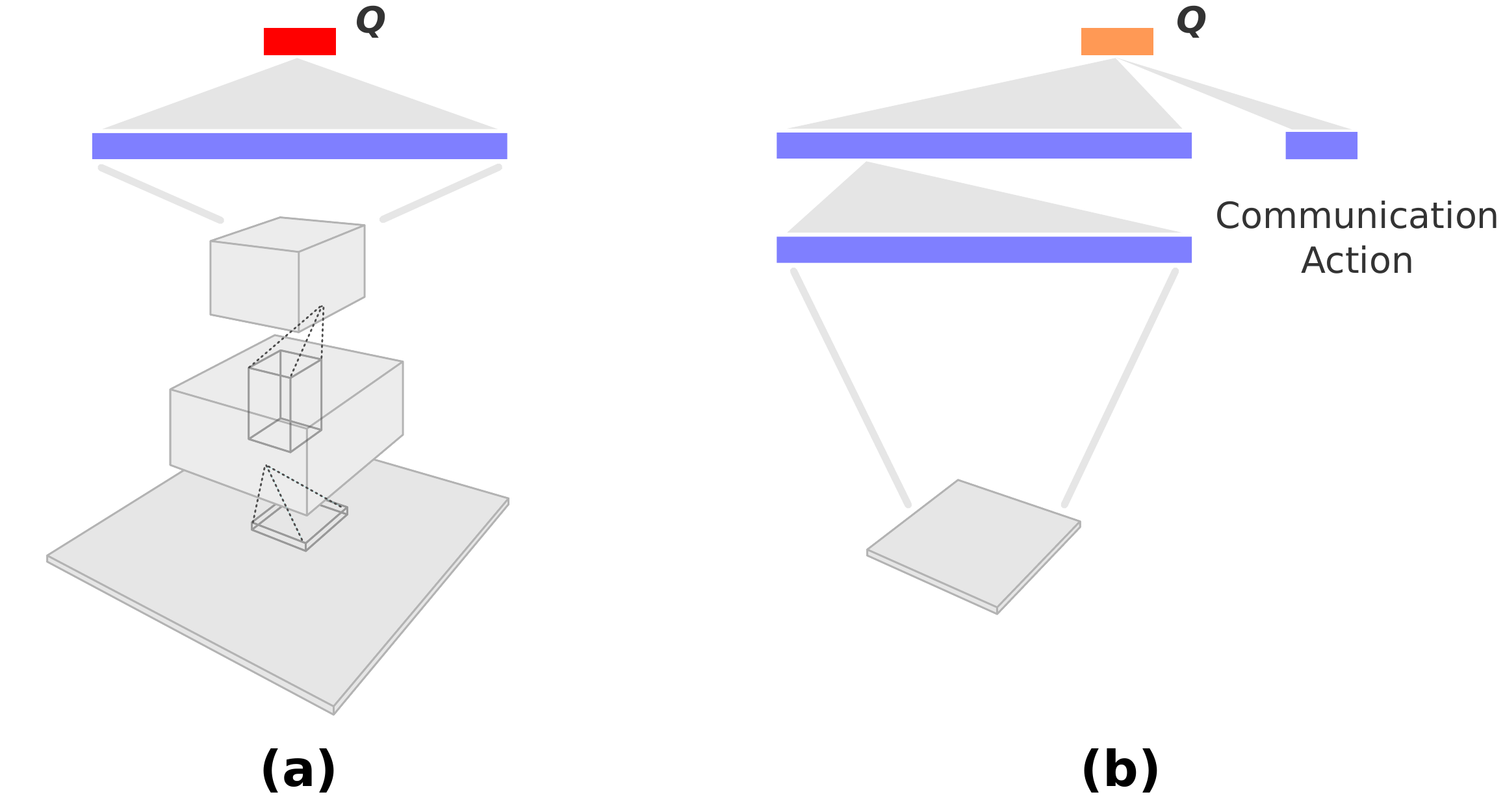}
\caption{ Illustration of the network used for the flat agent and the high-level agent {\it [a]} versus the network used for the low-level agent {\it [b]}. Because the low-level agent uses a bounding box, it does not require a full convolution network.}
\label{fig:catch_model}
\end{center}
\end{figure}

The low-level agent in the Catch experiment uses a Dense Network defined as follows. The input is passed through dense layers both containing 128 units and use rectified non-linear activations. The output of which is concatenated with the communication action sent by the high level agent; represented by a 1-hot vector of size $\vert Actions \vert = 3$. The merged representation is passed through the output layer with a linear activation and $\vert Actions \vert = 3$ units. A high-level illustration of the SoC system's architecture can be seen in Figure \ref{fig:catch_model}.

\end{document}